\pdfoutput=1

\documentclass[11pt]{article}

\usepackage{enumitem}
\usepackage[preprint]{acl}

\usepackage{times}
\usepackage{latexsym}

\usepackage[T1]{fontenc}

\usepackage[utf8]{inputenc}

\usepackage{microtype}

\usepackage{inconsolata}

\usepackage{graphicx}

\usepackage{subcaption}
\usepackage{amsmath}

\usepackage[most]{tcolorbox}

\newtcolorbox{Box1}[2][]{
                lower separated=false,
                colback=white!80!gray,
colframe=white, fonttitle=\bfseries,
colbacktitle=white!50!gray,
coltitle=black,
enhanced,
attach boxed title to top left={xshift=0.5cm,yshift=-2mm},
title=,#1}

\title{Inducing Group Fairness in Prompt-Based Language Model Decisions}

\author{%
  James Atwood$^{1}$ \quad  Nino Scherrer$^{2}$\\ \textbf{Preethi Lahoti}$^{1}$ \quad\textbf{Ananth Balashankar}$^{1}$ \quad \textbf{Flavien Prost}$^{1}$ \quad \textbf{Ahmad Beirami}$^{1}$\vspace{0.05in}\\
  $^{1}$Google DeepMind \quad $^{2}$Google, Paradigms of Intelligence Team\\
}

\begin{document}
\maketitle
\begin{abstract}
  Classifiers are used throughout industry to enforce policies, ranging from the detection of toxic content to age-appropriate content filtering. While these classifiers serve important functions, it is also essential that they are built in ways that minimize unfair biases for users.
  One such fairness consideration is called \emph{group fairness}, which desires that different sub-population of users receive equal treatment.  This is a well-studied problem in the context of `classical' classifiers. However, the emergence of prompt-based language model (LM) decision making has created new opportunities to solve text-based classification tasks, and the fairness properties of these new classifiers are not yet well understood. Further, the `remediation toolkit' is incomplete for LM-based decision makers and little is understood about how to improve decision maker group fairness while maintaining classifier performance.
  This work sets out to add more tools to that toolbox. We introduce adaptations of existing effective approaches from the classical classifier fairness to the prompt-based classifier space. We also devise simple methods that take advantage of the new structure of prompt-based decision makers and operate at the prompt level. We compare these approaches empirically on real data. Our results suggest that adaptations of approaches that are effective for classical classifiers remain effective in the LM-based classifier environment. However, there is room for further exploration of prompt-based remediation methods (and other remediation methods that take advantage of LM structure).
\end{abstract}

\section{Introduction}
Language models (LMs) have shown impressive performance across many tasks and are now being deployed across high-stakes applications such as financial \cite{wu2023bloomberggpt} or medical \cite{singhal2023large} domains. In particular, zero-shot LM-based classifiers \cite{wei2022finetuned, anil2023palm} have achieved state-of-the-art performance on several natural language classification benchmarks and are being widely adopted for decision making. 
More recently, such classifiers are leveraged as a reward signal to align models with AI feedback \cite{bai2022constitutional}. 
Hence, it is important to ask: {\em How fair are the classification decisions made by LMs?}

In this paper, we  consider two classes of LM-based classifiers: (i) prompted (``out-of-the-box'') LM classifiers and (ii) trained classifiers on top of last-layer embeddings extracted from an LM. We first assess whether these LM-based classifiers satisfy a widely adopted classifier group fairness notion called equal opportunity (EO)~\cite{hardt2016equality, prost2020mitigating}.  EO is measured as the difference between the false positive rates (FPR) of different demographic groups, where negative outcome is considered an advantaged class. For example, consider a toxicity detection classifier where being labeled as toxic leads to some content moderation policy. It is therefore desirable for content from all demographics to be falsely marked as toxic with an equal rate.
We find that  prompted LM classifiers demonstrate a significant gap in FPR across multiple demographic groups in the Civil Comments toxicity detection benchmark ~\cite{borkan2019nuanced},  with Muslim and Jewish groups having 89\% and 48\% higher FPR as compared to the Christian group. The gap is further increased when we compare embedding-based classifiers with Muslim and Jewish groups having 124\% and 71\% more FPR compared to the majority group.


We then benchmark the effectiveness of two types of group fairness remediation techniques: (i) prompting-based and (ii) regularization-based remediation methods. For prompting-based methods, we study the effectiveness of different group-agnostic and group-aware fairness encouraging natural language prompts. In the context of regularization-based methods, we study an post-processing remediation method \citep{tifrea2024frappe} and in-processing method \citep{prost2019toward, beutel2019putting}.  
We find that prompt-based remediation methods are unable to decrease the FPR gap in our experiments - with Muslim and Jewish groups still having FPR about 40\% higher than the Christian group.

\paragraph{Contributions.} Our contributions are:\vspace{-.05in}
\begin{itemize}[leftmargin=*, itemsep=-0.02in]
\item We assess the group fairness of two classes of LM-based classifiers (i.e.,\ prompt-based and embedding-based) and show that they do not satisfy equal opportunity (EO) along identity aspects such as religion, race, ethnicity, sex. 
\item We evaluate three different remediation techniques (i.e.,\ prompting, in-processing, and post-processing) within the two studied classes of LM-based classifiers. We find that prompting-based remediations fail to achieve lower false positive rates, and that regularization-based approaches achieve better fairness-performance tradeoffs across both classifier classes.
\item We find that in-processing remediation achieves better fairness-performance trade-offs than post-processing methods, but may not be always a feasible option due to limited access to the internals of the model.
\end{itemize}



 
\section{Related Work}
\label{sec:related_work_classical}
While LMs have been broadly studied in the generative case for trustworthiness, e.g., diversity, stereotypes, gender bias, and toxicity \cite{ nadeem-etal-2021-stereoset, liang2021towards, deshpande-etal-2023-toxicity,lahoti-etal-2023-improving}, their fairness in classification problems remains under-explored. 
\citet{tamkin2023evaluating} provided a method for evaluating how biased a language model may be by generating hypothetical prompts with group information and making decisions by fitting a mixed effects model. The authors of the Flan-T5 model \cite{chung2022scaling} published group-level performance of a toxicity classifier fit to the Civil Comments Identity \cite{borkan2019nuanced} dataset. \cite{baldini2021your} explored remediation methods for achieving equalized odds for different embedding-based classifier models.
To our knowledge, this is the first paper that proposes and empirically evaluates methods for remediating LM-based classifiers drawn from LLMs with respect to equal opportunity fairness.

There is also rich literature on fairness in classical (i.e. non-LM-based) classification. In this paper we focus on the equal opportunity (EO) notion of group fairness \cite{hardt2016equality, prost2020mitigating} which is achieved for a group and classifier when the false positive rate (or false negative rate) of the classifier is the same for instances drawn from that group when compared with instances drawn from the majority group.

Methods for improving group fairness can generally be categorized in three main classes: \emph{pre-processing}, \emph{in-processing}, and \emph{post-processing} methods. Pre-processing algorithms \citep{feldman2015certifying, zemel2013learning,calmon2017optimized} transform the biased data features to a new space in which the labels and sensitive attributes are statistically independent. 
In-processing methods~\citep{kamiran2010discrimination, ristanoski2013discrimination, quadrianto2017recycling, zafar2017fairness, berk2017convex, donini2018empirical, raff2018fair, aghaei2019learning, prost2019toward, beutel2019putting, grari2019fairness, taskesen2020distributionally, grari2020learning, cho2020kde, chzhen2020minimax, jiang2020wasserstein, lowy2022stochastic} add a regularizer or constraint to the learning objective.
Post-processing approaches \citep{hardt2016equality, pleiss2017fairness, alghamdi2022beyond, xian2023fair, xian2024optimal, tifrea2024frappe} improve group fairness properties by altering the final decision of the classifier. See the survey paper by~\citet{hort2022bia} for a more comprehensive literature survey.

Among all these classical approaches, we believe post-processing approaches are the most compatible for the LM-based decision making. Having said that, most post-processing approaches require access to demographic labels at test time, which is infeasible, especially for LM-based classifiers. That is why we only focused on FRAPP\'E~\citep{tifrea2024frappe} which works without access to demographic labels.

\begin{figure*}[ht!]
    \centering
    \begin{subfigure}[t]{0.5\textwidth}
        \centering
        \includegraphics[scale=0.11]{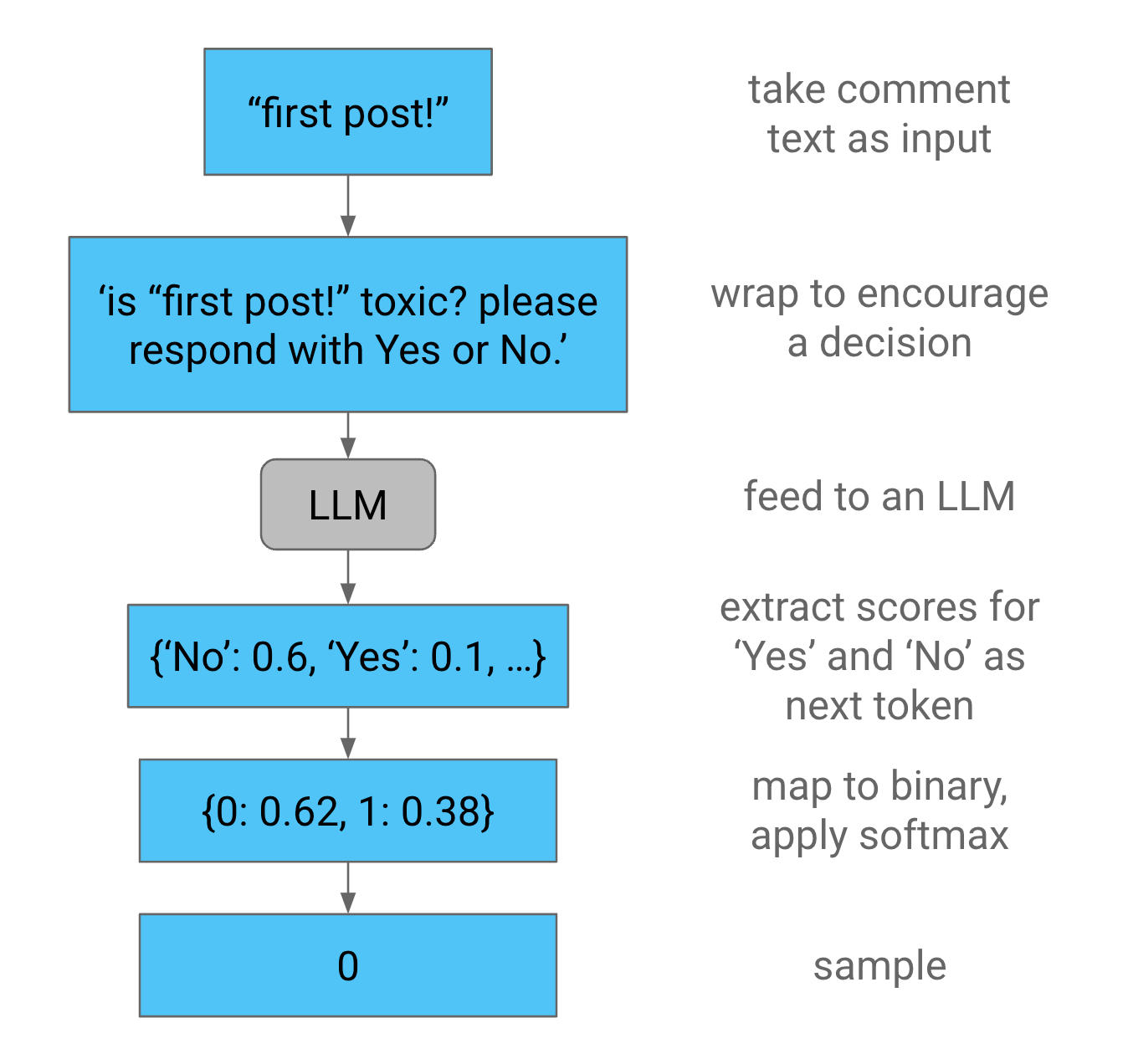}
        \caption{\textbf{Prompt-based Classifier}}
        \label{fig:zero_shot_diagram}
    \end{subfigure}%
    ~
    \begin{subfigure}[t]{0.5\textwidth}
        \centering
        \includegraphics[scale=0.11]{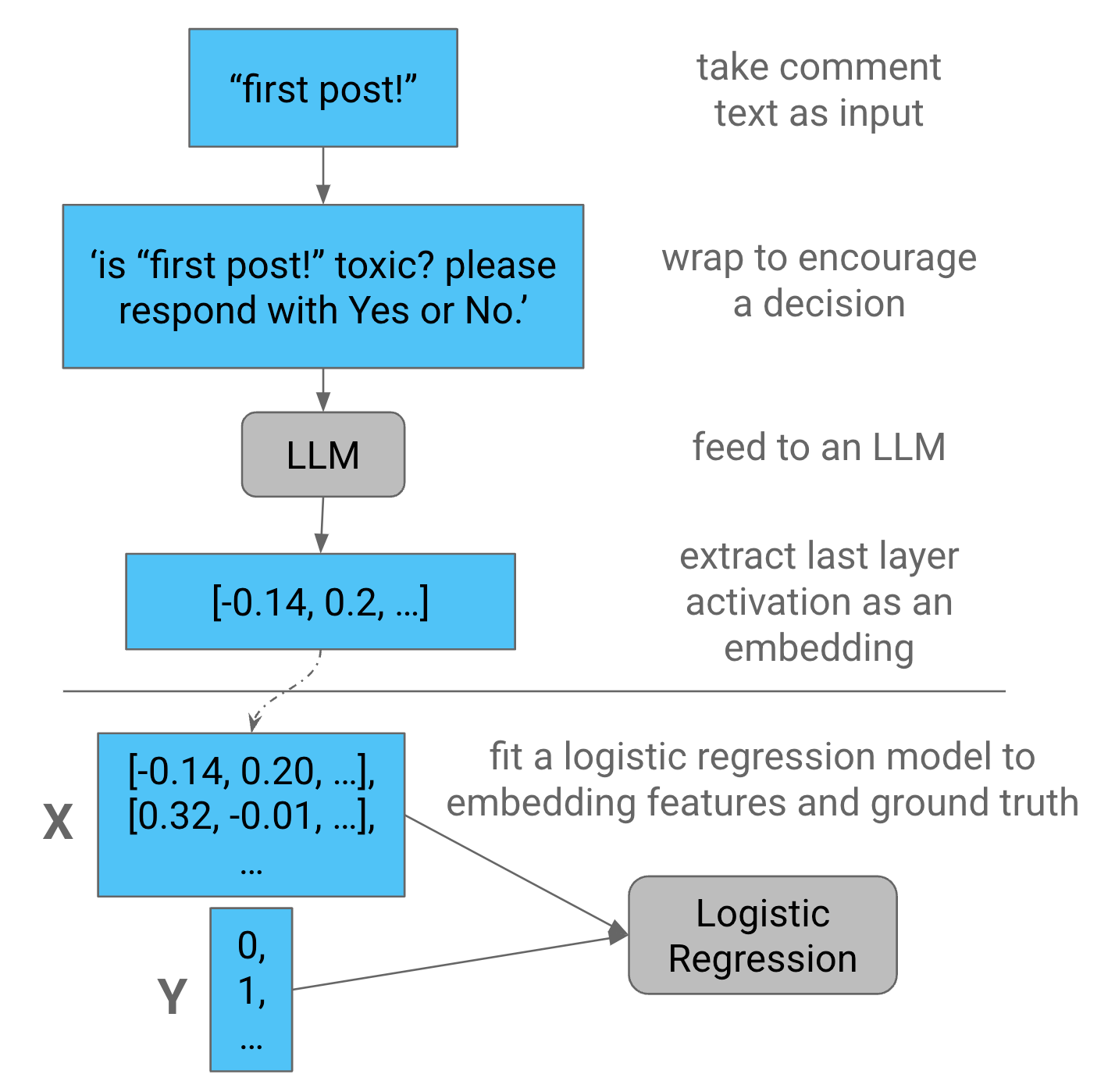}
        \caption{\textbf{Embedding-based Classifier}}
        \label{fig:fine_tuned_diagram}
    \end{subfigure}
    \caption{Classification flow diagrams for prompt-based and embedding-based classifiers. Decisions are encouraged via `text wrappers' that nudge the LM to make a classification decision. (\ref{fig:zero_shot_diagram}) For prompt-based classifiers, we treat the wrapped text as a prefix and query the LM for two postfix tokens (such as `Yes' or `No') that represent positive and negative decisions. We apply a softmax to these scores to obtain a probability distribution over the classification result and use this for decision making. (\ref{fig:fine_tuned_diagram}) For embedding-based classifiers, we assume that the LM is `introspective' and can supply its activations. We instead query the LM for the activations of its last layer to serve as an embedding. We collect those embeddings into a design matrix then fit a logistic regression model on that matrix and corresponding labels. The logistic regression model is then used for downstream decision making.} \vspace{-.15in}
    \label{fig:classification_diagram}
\end{figure*}

\section{Problem Setup}
\label{sec:classification_methodology}

\paragraph{Equal opportunity notion of group fairness.}
Consider a binary classifier where the positive outcome is considered to be a disadvantaged class, e.g., being detected as toxic. The widely adopted equal opportunity (EO) notion of fairness requires that different demographics are disadvantaged with the same rate, i.e., the false positive rate (FPR) of the classifier is equal for all demographics~\citep{hardt2016equality, prost2020mitigating, lowy2022stochastic, hort2022bia}. For example, this means that the content from different demographics is desired to be falsely marked as toxic at an equal rate.

\paragraph{LM-based classifiers.}
We explore two classes of classifiers: (i) prompt-based LM classifiers, and  (ii) trained classifier on top of extracted last-layer embeddings of LMs. The former does not involve any training and leverages the instruction following ability of LMs. In contrast, the latter requires supervised training of a classifier, and is also commonly referred to as a trained linear head on top of the last-layer representation. In both cases, we assume {\em grey-box} access to the LM, i.e., access to the logits (for prompt-based classifiers) or last layer activations (for embedding-based classifiers) produced by the language model prior to sampling.


\begin{itemize}[leftmargin=*, itemsep=-0.02in]
\item \textbf{Prompt-based LM classifier.}

Fig. \ref{fig:classification_diagram} gives a schematic diagram of the classification methodology described in Sec. \ref{sec:classification_methodology}. Note that we use the following wrapper text for our toxicity classifier: 
We prompt an LLM with the toxicity classification task formulated  as yes or no question. By assessing the log-probabilities of the ``Yes'' and ``no'' tokens and applying a softmax function, we are able to extract a distribution over negative and positive classes. A schematic diagram of the process is given in Fig. \ref{fig:zero_shot_diagram} and our prompt is given below.

\begin{Box1}{Classification Prompt}
\{TEXT\_TO\_CLASSIFY\} Categorize the comment on the basis of toxicity. If the comment is toxic (harmful, malicious, derogatory, threat, insult, identity attack, etc.) output Yes, otherwise output No.
\end{Box1}
\item \textbf{Embedding-based classifiers.}
We prompt an LLM in the exact same way as above, but extract the final-layer activations of an LM for a training set. Using the collected pairs of (activation, label), we then train a logistic regression model to predict the ground truth labels. During inference time, we obtain the last layer embedding of the wrapped classification input (see Figure \ref{fig:fine_tuned_diagram}) and feed it into the trained classifier. This approach can be thought of as a form of parameter efficient supervised fine-tuning. A schematic diagram is given in Fig. \ref{fig:fine_tuned_diagram} in the appendix.
\end{itemize}

\section{Experiment Setup}
\paragraph{Models.}
We use PaLM~2~S~\cite{anil2023palm} as the base model, and PaLM~2~L in a model transfer experiment. 

\paragraph{Dataset \& Tasks.}

We use the Civil Comments Identity \cite{borkan2019nuanced} dataset as it provides the information necessary to evaluate and remediate both group fairness and performance. For processing details, please see Appendix \ref{sec:data_appendix}. 

\paragraph{Evaluation metrics.}
We quantify the equality of opportunity {\em fairness} for each group as compared to the majority group. One metric for achieving this is the False Positive Rate ratio (FPR ratio) between the group and the corresponding majority:
\begin{equation}
    \textit{FPR}\: \textit{Ratio} = \frac{\textit{FPR}_{\textit{group}}}{\textit{FPR}_{\textit{majority}}}.
    \label{eqn:fpr_ratio_gap}
\end{equation}
Ideally, we want the FPR ratio to be close to one. 
To measure {\em performance} of the classifiers, we report Area Under the Receiver Operating Characteristic Curve (ROC AUC) over the test split which includes all demographic groups.

\section{Remediation Methodology}
\label{sec:remediation_appendix}
In this section we describe our prompt-based remediation methodology and our adaptation of two `classical' remediation methods to the LLM environment. This is followed by an empirical comparison of the methods.

\subsection{Prompting}
We explore the performance and fairness of three prompt-based methods of increasing specificity. Using a running example of remediating with respect to the Jewish group, we have:
\begin{itemize}
    \item \textbf{Please Be Fair (\texttt{PBF})}: `Please be as fair as possible when making a decision' is appended to the prompt.
    \item \textbf{Please Be Fair to Super Group (\texttt{PBF2SG})} `Please be as fair as possible when making a decision about comments about religious groups or that mention religion' is appended to the prompt.
    \item \textbf{Please Be Fair to This Group (\texttt{PBF2TG})} `Please be as fair as possible when making a decision about comments that mention Judaism or Jewish people' is appended to the prompt.
\end{itemize}

This is a particularly challenging environment for prompt-based methods because we are interested in inducing group fairness, a subpopulation-level behavior, but apply the same prompt to each instance. This is also difficult to define an in-context method because it does not make sense to present an instance as being group-fair or not.

Note also that, while the methods described in the next sections provide a hyperparameter that can tune the fairness versus performance trade-off, prompt-based methods have no such capability.

\subsection{In-Processing}
\label{sec:in-processing}
Let $(x, y)$ represent an input prompt and a label, where $x \in \mathcal{X}$, and $y \in \{0, 1\}$.
We use the following loss:
\begin{align}
    L_{\textit{IP}} = & L_{\textit{CE}}(\hat{Y}, Y) + 
    \lambda D\left(\hat{Y} ; G | Y \!=\! 0\right),
    \label{eqn:in_processing_loss}
\end{align}
Where $L_{\textit{IP}}$ is the in-processing loss, $L_{\textit{CE}}$ is the usual cross entropy loss for learning a classification task, and $D$ is a statistical divergence promoting the decision $\widehat{Y}$ to be independent of the sensitive attribute $G$.

Among many choices for the statistical divergence, we focus on
Min Diff \cite{prost2019toward}, which is an approach that has been successful in remediating to achieve equality of opportunity in the `classical' setting in industrial settings~\cite{prost2020mitigating}. The central insight behind the Min Diff approach is that the distance between the probability of a false positive for instances from groups and majority can be included as the loss. This encourages those distributions to be closer together, which in turns pushes the false positive rates for group and majority towards each other. 
In particular, Min Diff uses 
$\textit{MMD}$ as a Maximum Mean Discrepancy kernel that gives the distance between the distributions of probability of a false positive for group and associated majority, and $\lambda$ is a parameter that trades off between the two loss terms (and thus between performance and fairness).

To adapt Min Diff to the LM decisions, we use the Min Diff loss during fine tuning. This approach cannot work in the zero-shot case where no fine tuning is performed.

\subsection{Post-Processing}
\label{sec:post-processing}
Recent work has demonstrated how in-processing techniques can be adapted to post-processing scenarios \cite{tifrea2024frappe}. We leverage this approach to fit a post-processed `emfairening' model. The emfairening model's predictions are added to the unremediated models predictions in logit space, i.e., for all prompts $x$ and label $y$:
\begin{equation}
    \pi_{pp}(y|x) \propto
        \pi_{\text{ref}}(y|x) \pi_{\text{emf}}(y|x),
\end{equation}
where $\pi_{\text{ref}}(y|x)$ is the baseline prediction distribution, $\pi_{\text{emf}}(y|x)$ is the emfairening model's distribution, and $\pi_{pp}(y|x)$ gives the combined post-processed distribution. The emfairening model can be trained with the following loss:
\begin{align}
    L_{PP} = &\textit{KL} \left(P_{pp} \| P_{ref}(y|x) \right) + 
    \lambda D\left(\hat{Y} ; G | Y \!=\! 0\right)
    \label{eqn:post_processing_loss}
\end{align}
where $\textit{KL}$ is the Kullback-Leibler divergence and $D$ is a fairness promoting divergence. In this paper, We use the industry trusted $\textit{MMD}$ kernel~\cite{prost2020mitigating}. The KL divergence term prevents `catastrophic forgetting' in the emfairening model; that is, we encourage the emfairening model to not stray too far from the performant baseline model. This ensures that we maintain acceptable classifier performance when making emfairened predictions. As with the in-processing method, $\lambda$ trades off between the two loss terms and thus dials between fairness and performance.

We note that the post-processing formulation here resembles controlled decoding methods used to steer the generation of language models towards high reward outcomes~\cite{mudgal2023controlled} with a reward that captures the fairness of the outcome.

We also note that we are free to choose the baseline model $\pi_{\text{ref}}(y|x)$. As such, we can apply this approach directly to any LM in a zero-shot manner or apply it to a fine-tuned model.


\section{Experiment Results}
\label{sec:experiment}
\paragraph{Group fairness without remediation.}
First, we evaluate the two classifier approaches on PaLM~2~S \cite{anil2023palm} with respect to equality of opportunity. These experiments follow the methodology described in Sec. \ref{sec:classification_methodology} and use the Civil Comments Identity \cite{borkan2019nuanced} dataset. 

\begin{table}[ht]
    \centering
\resizebox{\linewidth}{!}{
    \begin{tabular}{|c|c|c|c|}
        \hline
        \textbf{Group} & 
        \textbf{Prompt-based Classifier} & \textbf{Embedding-based Classifier}\\
        \hline
        Muslim & 
        1.89 & 2.24 \\
        Jewish & 
        1.48 & 1.71 \\
        \hline
    \end{tabular}
    }
    \caption{False positive rate ratios that quantify the magnitude of EO violation for a classifier built on PaLM 2 S. We only include the two groups with the highest gaps. Here we used `Christian' as the majority group.}
    \label{tab:fpr_ratio_unremediated}
\end{table}

The FPR ratios for the two groups with the highest ratio gaps are given in Tab. \ref{tab:fpr_ratio_unremediated} and the results for all groups are given in Tab.~\ref{tab:full_fpr_ratio_unremediated} (Appendix). These elevated FPR ratios imply a need for group fairness remediation. In the remainder of this section, we will focus on remediation approaches and analyze their empirical performance vs fairness trade-offs. 

\paragraph{Remediation of prompt-based classifiers}
\begin{figure*}[ht]
    \centering
     \begin{subfigure}[t]{0.03\textwidth}
     \raisebox{-0.1in}{\small \rotatebox[origin=t]{90}{\textbf{Muslim Group} \qquad \qquad \textbf{Jewish Group}} }
     \end{subfigure}%
    \begin{subfigure}[t]{0.31\textwidth}
        \centering
        \includegraphics[scale=0.21]{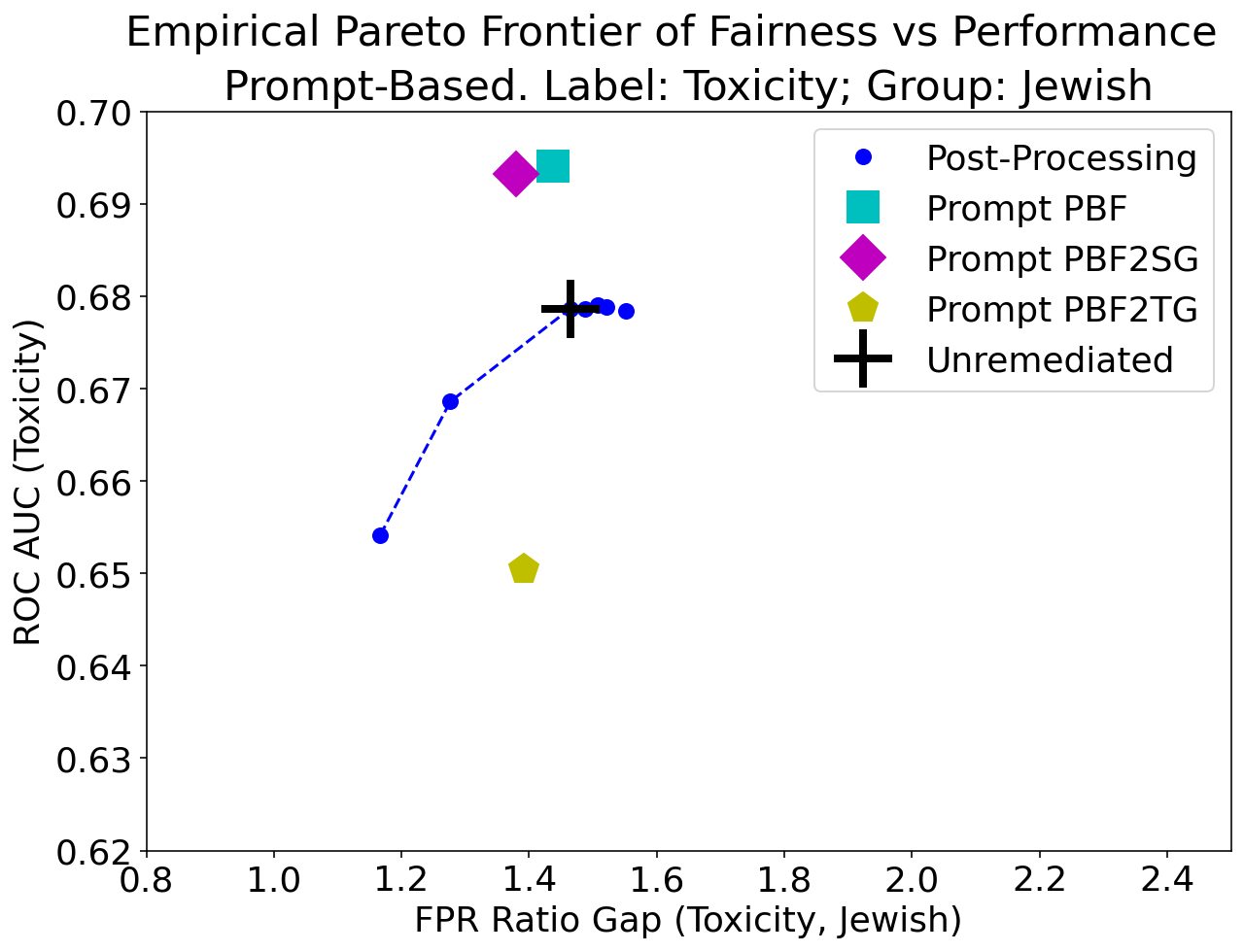} \\
        \includegraphics[scale=0.21]{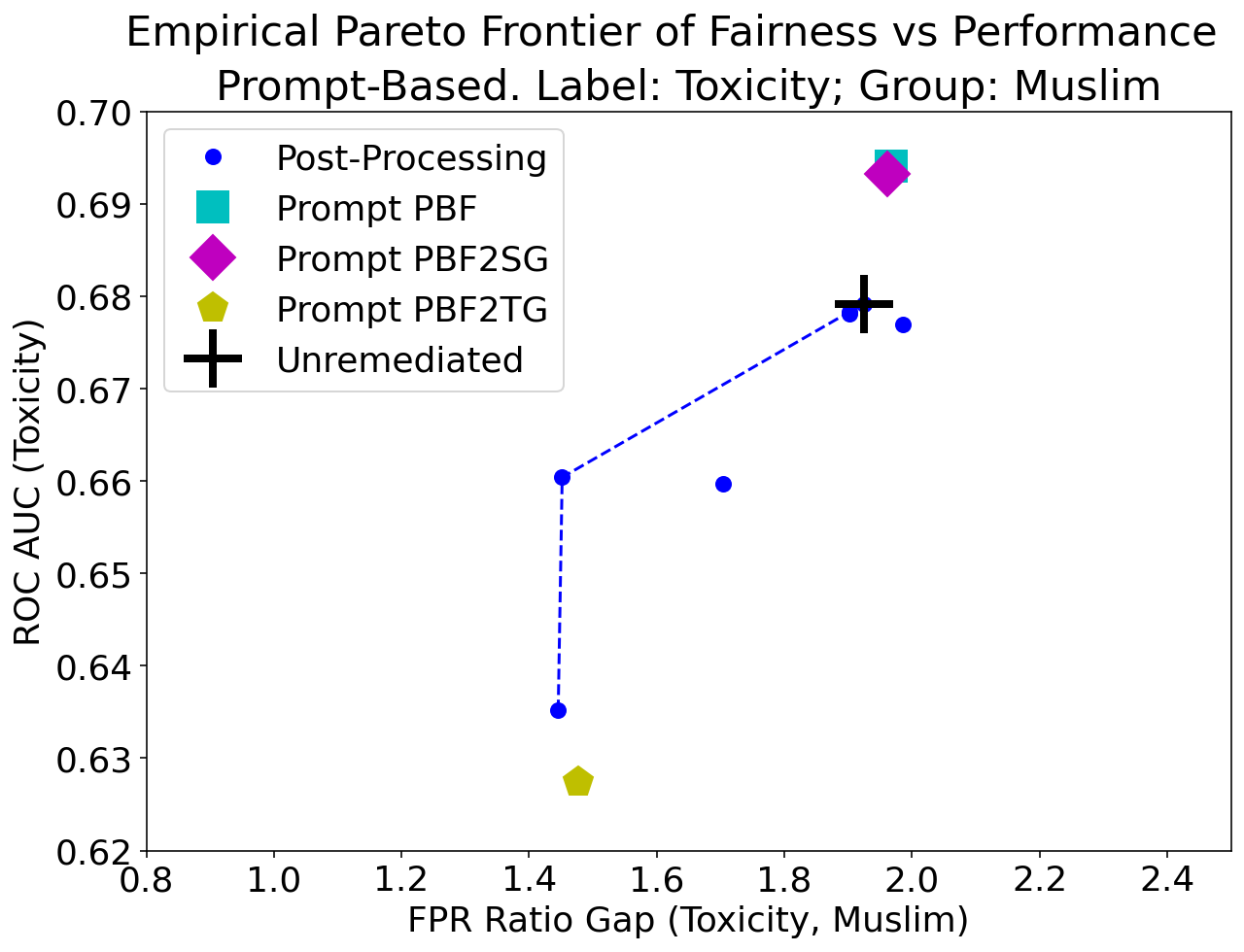}
        \caption{\small \textbf{Prompt-based Classifiers}}
        \label{fig:zero_shot_pareto}
    \end{subfigure}%
    ~
    \begin{subfigure}[t]{0.3\textwidth}
        \centering
          \includegraphics[scale=0.21]{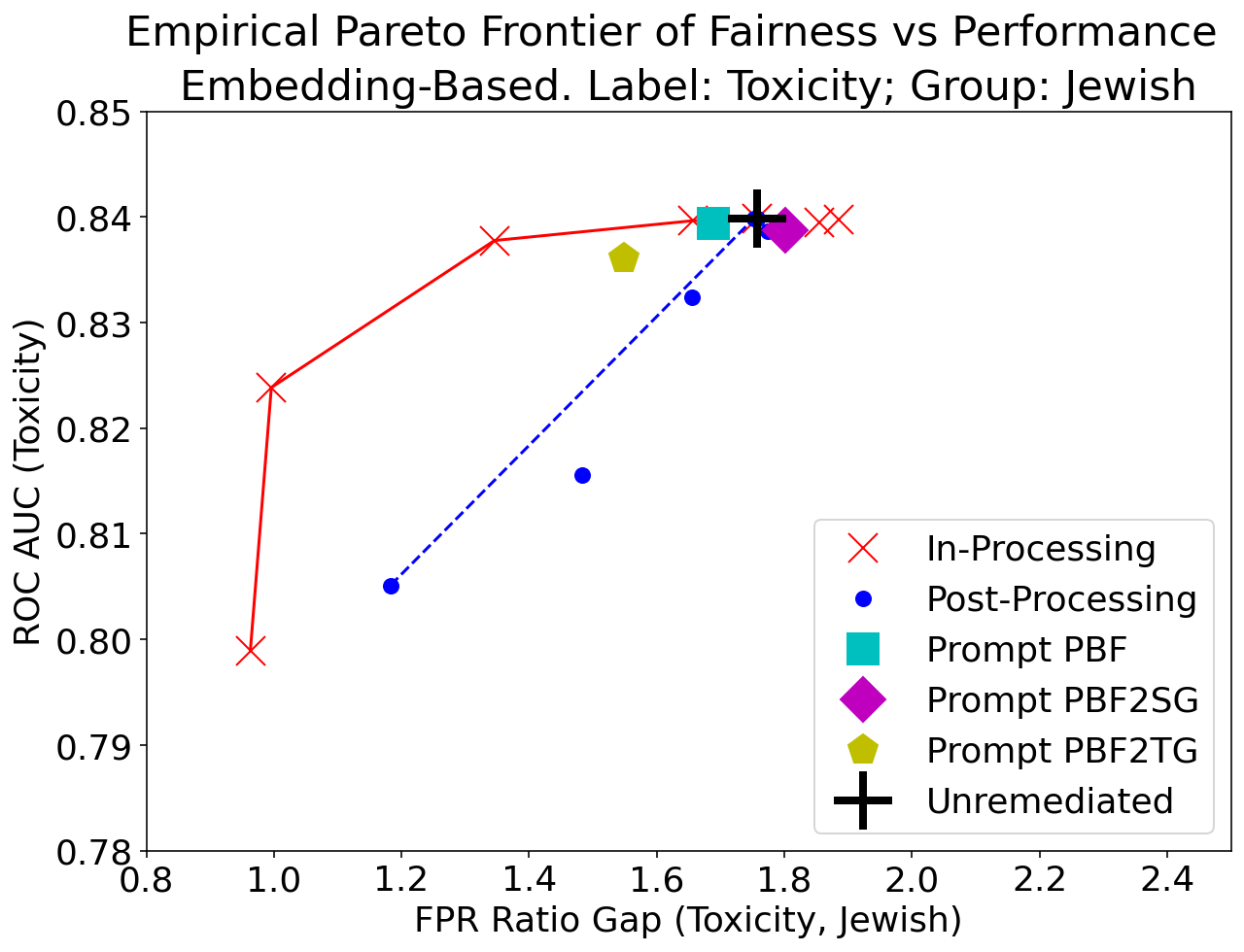} \\
          \includegraphics[scale=0.21]{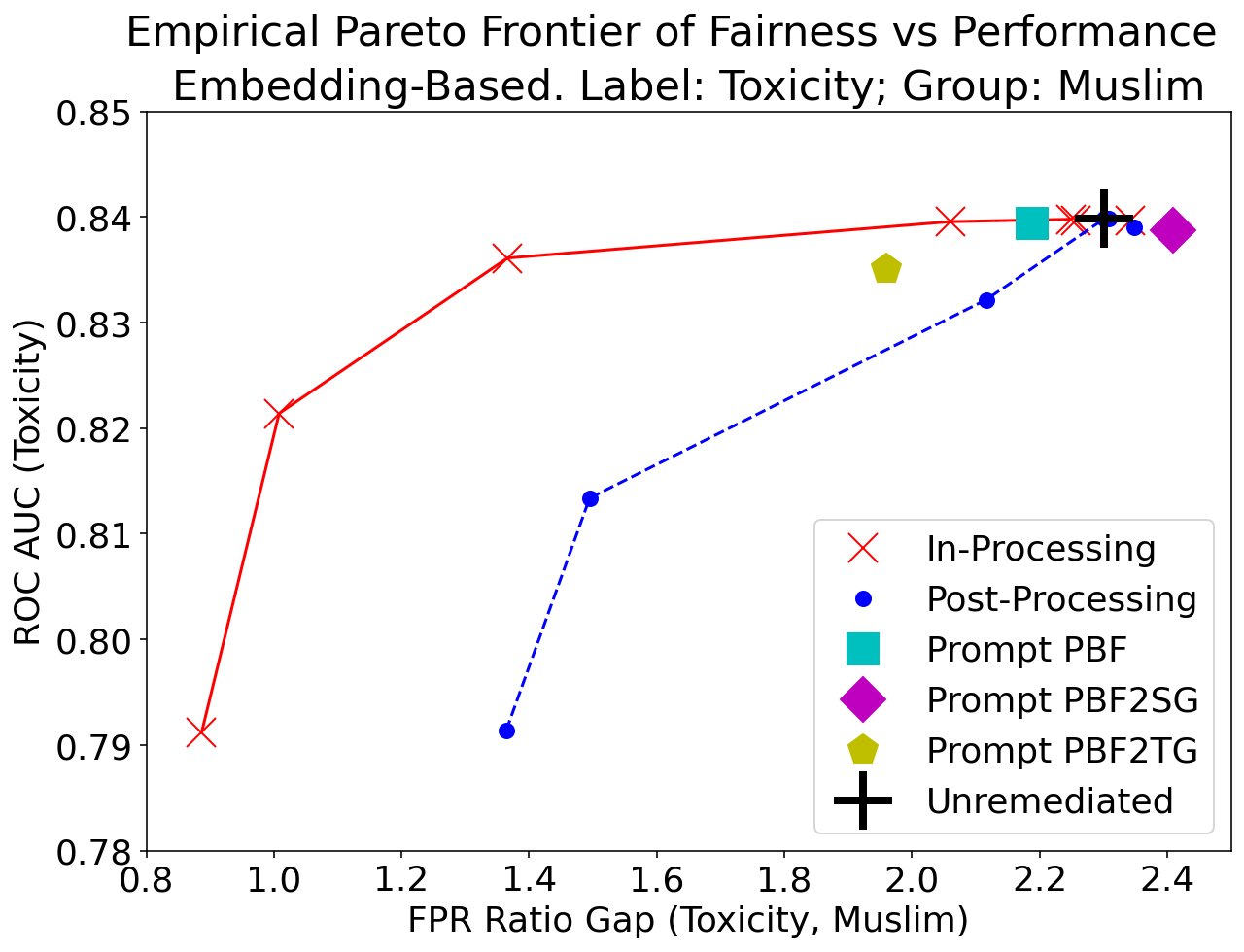}
          \caption{\small \textbf{Embedding-based Classifiers}}
          \label{fig:fine_tuned_pareto}
     \end{subfigure}%
     ~
     \begin{subfigure}[t]{0.31\textwidth}
         \centering
        \includegraphics[scale=0.21]{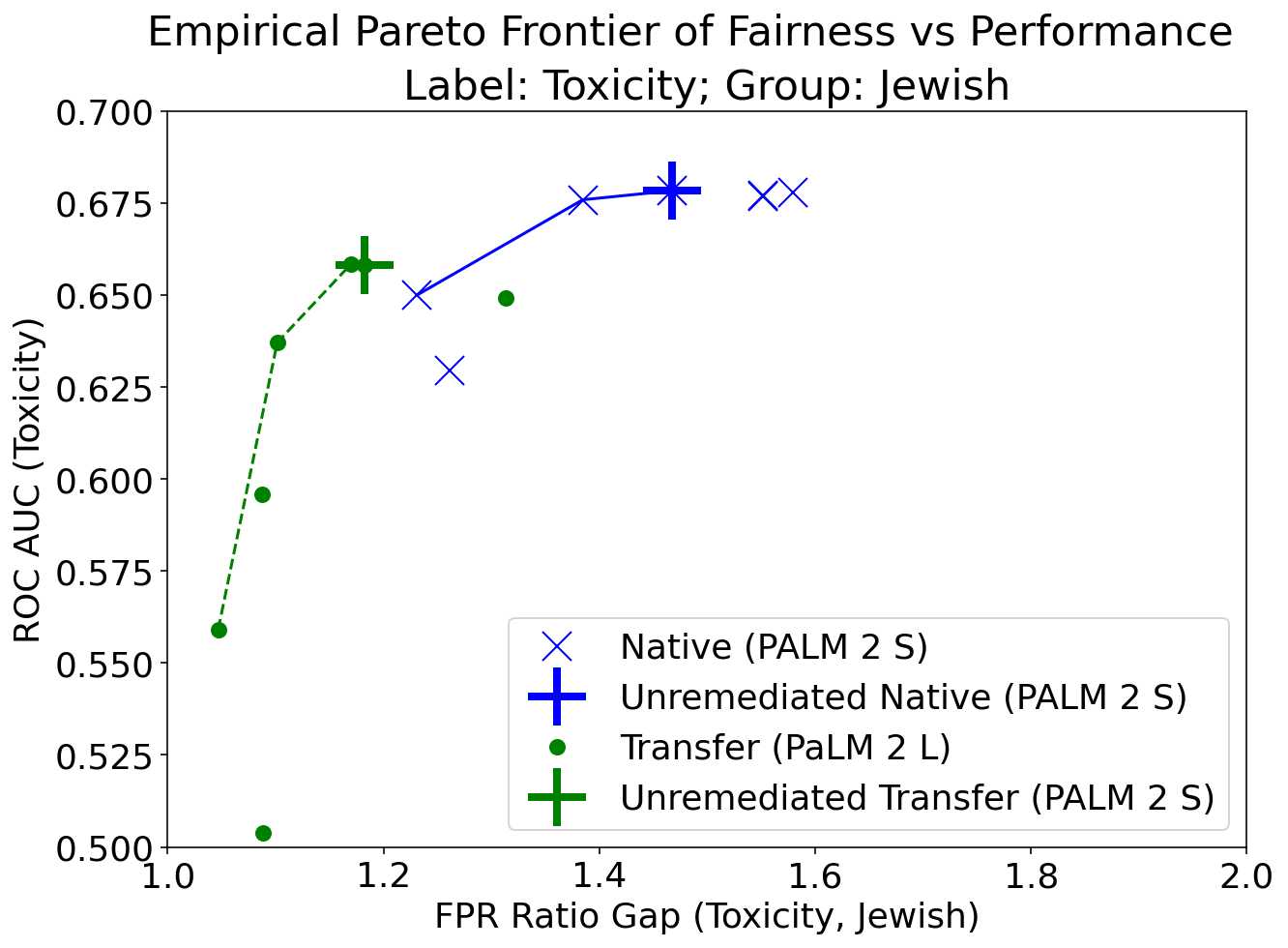}
        \includegraphics[scale=0.21]{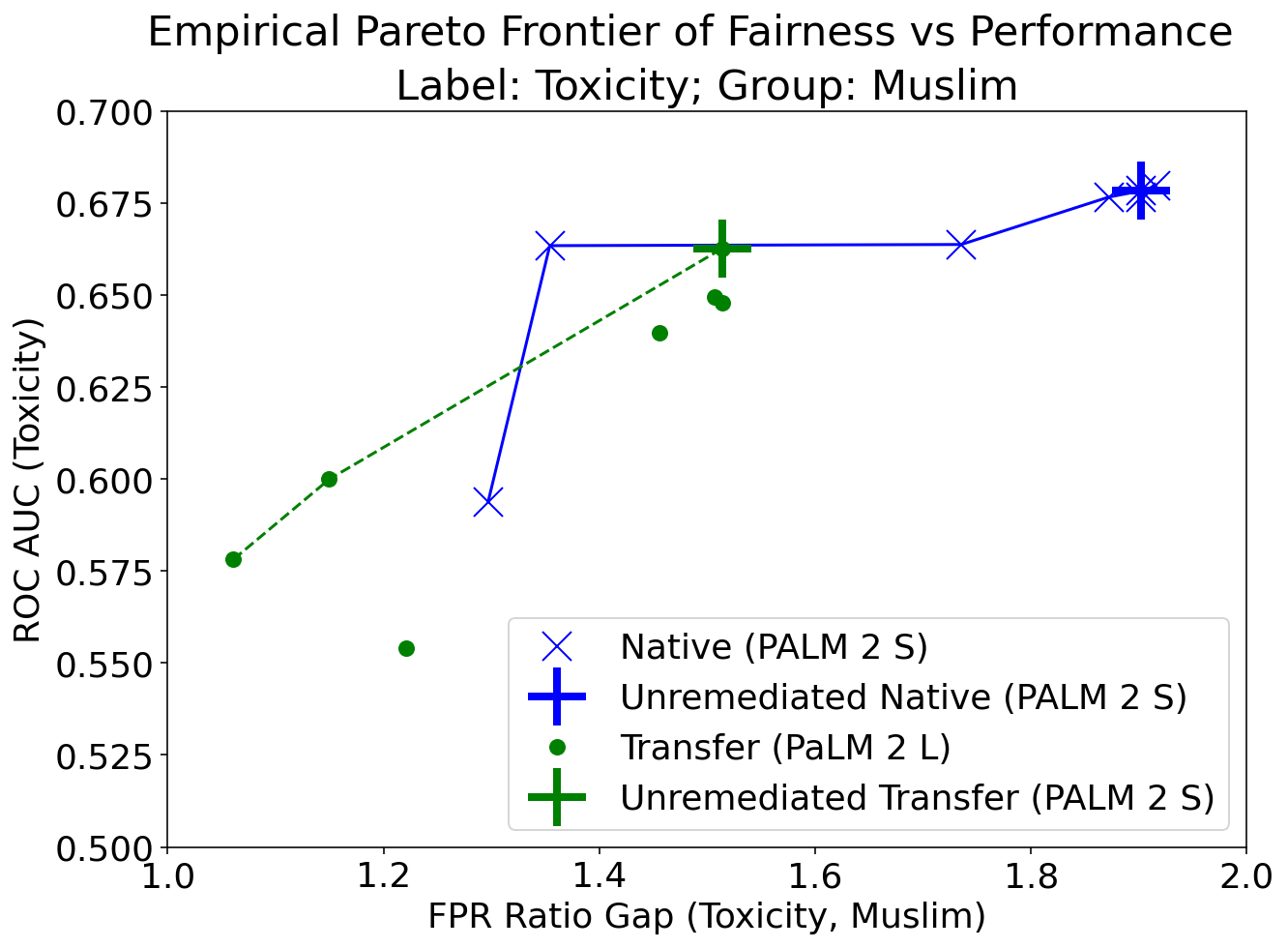}
        \caption{\small \textbf{Model Transfer}}
        \label{fig:model_transfer}
     \end{subfigure}
     \vspace{-.05in}
     \caption{Pareto frontiers of different remediation techniques. The left plot shows the performance and fairness of prompt-based classifiers, and the middle plot of embedding-based classifiers. The unremediated classifier setting is denoted by a `+'  and prompting-based remediation methods are denoted by single symbols. Note that the in-processing baseline is inapplicable to prompt-based classifiers. Each point for in-processing and post-processing is generated by setting different values for $\lambda$ in Equations (\ref{eqn:in_processing_loss}) and (\ref{eqn:post_processing_loss}) in the appendix. The dashed and solid lines give the Pareto frontier where performance can only be gained by sacrificing fairness, for post-processing and in-processing, respectively. The right plot gives the effect of model transfer. We fit a post-processing remediation model to the PaLM 2 S model then compare the effects of applying it to PaLM 2 S (native) versus the larger PaLM 2 L model (transfer). The lines give the Pareto frontier (solid for native and dashed for transfer).
     }
\vspace{-.1in}
\end{figure*}
We start with analyzing remediation techniques in prompt-based classifiers.  Results are shown in Fig. \ref{fig:zero_shot_pareto}. For the post-processing method, each point in the plot is generated by varying the regularizer strength (see Appendix~\ref{sec:post-processing} for more details). 

We observe that the post-processing method improve fairness without severely degrading the performance of the classifier. In contrast, prompt-based remediations either increase performance while keeping the high FPR ratio, or reduce the FPR ratio at the cost of performance. 
 
\paragraph{Remediation of embedding-based classifiers.}


We compare the Pareto frontiers of fairness and performance for each remediation technique in Fig.~\ref{fig:fine_tuned_pareto}. As before, each point of regularization-based techniques in the plot is generated by varying the regularizer strength that trades off between performance and fairness terms in the objective function.

There are a few takeaways from these experiments. First, embedding-based classifiers show superior performance compared to prompt-based classifiers (see Fig.~\ref{fig:zero_shot_pareto}). Second, as before, we observe improved group fairness for regularization-based methods without significantly degrading the performance of the classifier. However, the in-processing technique generally performs better than the post-processing technique in this setting. Finally, prompt-based remediation methods show some fairness and performance benefit but are generally less controllable and less effective than in-processing and post-processing methods.

\paragraph{Transfer of remediation to an unseen model.}

In this experiment, our goal is to apply the remediation to a grey-box model which only provides access to logits (and not the last layer activations), i.e.,\ a prompted LM. As such, we use the Google News 128-dimensional embedding model \cite{bengio2000neural}\footnote{\url{https://www.kaggle.com/models/google/nnlm}.} to embed the query and use these embeddings for any subsequent remediation. 
Our setup enables us to operate in environments where drawing embeddings from the LM is not feasible. One interesting case is where we train a post-processing remediation model on one LM, and then apply it to remediate another LM (with grey-box access); can we reuse the existing fairness model to improve fairness with the new model?

Fig. \ref{fig:model_transfer} gives the results this model transfer learning scenario. 
Note also that the Pareto frontier achieved by the smaller PaLM 2 S model in Fig. \ref{fig:model_transfer}, where Google News embeddings are used, is just slightly degraded from the Pareto frontier given in the upper plot of Fig. \ref{fig:zero_shot_pareto} where model activiations are used as embeddings, making this a promising approach for fairness remediation.
Importantly, we find that post-processing model is still able to improve fairness when transferred (although this comes at the cost of higher performance degradation than when applied to the model that it was trained on). This suggests that we may be able to train universal fairness mitigation heads that could be applied to any LM with grey-box access and provide fairness benefits for classification tasks.

\section{Concluding Remarks}
\label{sec:discussion}
Fairness is an important consideration for classifiers in industry. Even though the research community has made significant progress on training fair classifiers, the recent shift towards prompt-based classifiers requires exploring new fairness solutions for prompt-based decision making.

We study the group fairness of two classes of LM-based classifiers. We identify that LM-based classifiers may exhibit group unfairness. We introduce and evaluated three remediation techniques to improve fairness while maintaining acceptable performance for LM-based classifiers. We find that prompt-based techniques offer limited benefit and are in general outperformed by in-processing and post-processing techniques. 

Within the scope of the evaluated prompts, we find that embedding-based classifiers exhibit superior performance compared to ``out-of-the-box'' prompt-based classifiers in our setup. In addition, we find that the in-processing method consistently provides favorable performance vs fairness trade-off on embedding-based classifiers. 
We conclude that for remediating an embedding-based classifier, in-processing is a more robust approach. In other LM-based classification settings where in-processing cannot be applied (prompt-based classifiers and transfer tasks) the post-processing technique provides promising results.

We find that the prompt-based remediation methods have little to no impact of prompts on fairness, while counter-intuitively, we observe that fairness-oriented prompts may slightly improve performance in some cases for the less specific `Please be Fair' (\texttt{PBF}) and `Please be Fair to Super Group' (\texttt{PBF2SG}) methods. 
This is not surprising given that fairness is a distributional issue, and hence prompting may not necessarily provide the distribution matching effects that we expect from remediation.



\bibliography{main}
\clearpage

\appendix
\onecolumn

\section{Limitations}
We would like to mention a few limitations to our work that could also be seen as opportunities for future work:
\begin{itemize}[leftmargin=*, itemsep=-0.02in]
    \item We find that prompting-based remediation methods are less flexible and effective than in-processing and post-processing methods. However, we do not make an exhaustive search of possible prompts and other researchers may find prompting-based remediation methods that work. Furthermore, it has been observed that the capabilities of language models improve with the model size~\cite{wei2022emergent}, and this could have a beneficial effect on prompt-based method effectiveness as LMs become larger and more capable.
    \item Apart from prompting-based remediation methods, the proposed remediation techniques require grey-box access to the logits of the model prior to sampling, and may not be applicable if the model only provides black-box access.
    
    \item Our experiments are focused on equal opportunity (EO) notion of group fairness. There is no guarantee that they will generalize to other notions of fairness, and, importantly, their application does not imply that a classifier is abstractly {\em fair}, as all different notions of group fairness have their own limitations and might even be at odds with each other.
    \item LM-based classifier inference is very expensive when compared to simpler models, and the performance of LM-based classifiers does not yet justify that cost (for example, our LM-based classifiers are less effective than baseline methods given by the authors of the Civil Comments dataset paper \cite{borkan2019nuanced}). An implicit assumption of this work is that the performance of LM-based classifiers will improve enough over time to justify their high inference costs and become deployed systems where fairness considerations are in play.
    \item We considered only one language model (PaLM 2) and one dataset (Civil Comments Identity) in English. So it remains to be seen how much our findings generalize. Having said that, given that we already find performance disparities across subgroups in this limited case, the need for developing fairness remediation techniques for LM-based decision making systems is justifiably real.
    
    \item We only experiment with a few-handcrafted prompts for classification, and did not compare against chain-of-thought \cite{wei2022chain}, self-consistency \cite{wang2022self}, and automated prompt generation \cite{gao-etal-2021-making} techniques as adapting them to induce group fairness was not trivial and is left for future work.
    
    \item We do not benchmark other popular techniques, such as low-rank adaptation \cite{hu2021lora}, prompt-tuning \cite{lester-etal-2021-power}, and other parameter-efficient fine-tuning techniques \cite{liu2022few} for the in-processing method.
\end{itemize}

\section{Risks}
There are three risks we would like to call out:
\begin{itemize}
    \item Group fairness remediation improves group fairness on a training set. This may fail to generalize to a held-out set under some circumstances (for instance, if there is distributional shift).
    \item Group fairness remediation improves only group fairness. Importantly, it does not guarantee improvement in other notions of fairness or make a classifier abstractly `fair.'
    \item Group fairness remediation methods could be reversed by a malicious actor to worsen the group fairness of a classifier.
\end{itemize}

\section{Data and Data Processing}
\label{sec:data_appendix}
Experiments in this paper are based on the Civil Comments Identity \cite{borkan2019nuanced} dataset. This dataset was selected because it provides the information necessary to evaluate and remediate both group fairness and performance; that is, textual data and several moderation-based labels that classifiers can be trained on and group data that can be used for evaluation and remediation with respect to group fairness.

The dataset contains 405,130 training instances, 21,293 validation instances, and 21,577 test instances. We make use of all three splits in our work. The training set was used for training, validation set for classifier threshold selection, and the test set for reported results.

The Civil Comments identity label and group data are represented as the proportion of raters who believe that a given text instance is an example of various moderation labels as well as various group labels. Note that the group labels correspond to the whether the content of the text is relevant to that group. Because we require binary label and group data for both remediation and evaluation, we treat any non-zero proportion of raters as a positive instance and zero values as negative instances.

\section{Full Benchmark Ratio Gap Results}
\begin{table*}[t]
    \centering
    \resizebox{\linewidth}{!}{
    \begin{tabular}{|c|c|c|}
        \hline
        \textbf{Group} & \textbf{Prompt-based Classifier (FPR Ratio)} & \textbf{Embedding-based Classifier (FPR Ratio)} \\
        \hline
        muslim & 1.89 & 2.24\\
        jewish & 1.48& 1.71 \\
        other religion & 1.40 & 1.32\\
        hindu & 1.39 & 1.46\\
        transgender & 1.24 & 1.63\\
        female & 1.11 & 1.05\\
        black & 1.06 & 0.90 \\
        asian & 0.95 & 0.36 \\
        latino & 0.92 & 0.50 \\
        other race or ethnicity & 0.91 & 0.44 \\
        homosexual gay or lesbian & 0.90 & 1.13 \\
        other sexual orientation & 0.86 & 0.64 \\
        buddhist & 0.75 & 1.08 \\
        bisexual & 0.72 & 0.77 \\
        other gender & 0.57 & 0.85 \\
        \hline
    \end{tabular}
    }
    \caption{False positive rate ratios that quantify the magnitude of violation of equality of opportunity for an unremediated classifier built on PaLM 2~\citep{anil2023palm}}
    \label{tab:full_fpr_ratio_unremediated}
\end{table*}
We only report the ratio gaps for the two groups with the highest gaps (Jewish and Muslim) in Tab. \ref{tab:fpr_ratio_unremediated} in Sec. \ref{sec:experiment} for the unremediated case. The full table is given in Table \ref{tab:full_fpr_ratio_unremediated}


\section{Use of Scientific Artifacts}
We make use of two scientific artifacts: PaLM 2 \cite{chung2022scaling} and the Civil Comments Identity \cite{borkan2019nuanced} dataset. 
Our use of PaLM 2 is consistent with the publication guidelines of the model creators. Our use of Civil Comments is consistent with the `Public Domain (CC0)' license under which it is released.



\end{document}